\documentclass[11pt]{article}

\usepackage[margin=1in]{geometry}
\usepackage{times}
\usepackage{microtype}
\usepackage{amsmath,amssymb}
\usepackage{graphicx}
\usepackage{booktabs}
\usepackage{tabularx}
\usepackage{array}
\usepackage{multirow}
\usepackage{xcolor}
\usepackage{hyperref}
\usepackage{url}
\usepackage{enumitem}
\usepackage{listings}
\usepackage{tikz}
\usetikzlibrary{arrows.meta, positioning, shapes.geometric, fit, backgrounds}
\usepackage{natbib}
\usepackage{authblk}
\usepackage{abstract}
\usepackage{titlesec}
\usepackage{caption}
\usepackage{subcaption}

\hypersetup{
  colorlinks=true,
  linkcolor=blue!70!black,
  citecolor=blue!70!black,
  urlcolor=blue!70!black,
}

\title{
  \textbf{PRISM: Prompt Reliability via Iterative Simulation\\
  and Monitoring for Enterprise Conversational AI}
}

\author[1]{Keshava Chaitanya}
\author[1]{Jahnavi Gundakaram}
\affil[1]{Yellow.ai \quad \texttt{\{keshava, jahnavi\}@yellow.ai}}

\date{}

\begin{document}

\maketitle

\begin{abstract}
Deploying large language model (LLM)-driven conversational agents in enterprise
settings requires prompts that are simultaneously correct at launch and resilient
to the non-deterministic behavioral drift that characterizes production LLM
deployments. Existing prompt optimization frameworks address prompt quality as a
one-time compile-time problem, leaving open the equally critical question of how
to detect and repair prompt regressions caused by silent LLM behavior changes
over time. We present \textbf{PRISM} (Prompt Reliability via Iterative
Simulation and Monitoring), a closed-loop framework that treats prompt
engineering as a continuous reliability engineering problem rather than a
one-time authorship task. PRISM takes as input plain-language agent requirements,
a set of configured tools and memory variables, and an initial draft prompt. It
automatically generates test cases from requirements, simulates full multi-turn
conversations against a platform-faithful LLM environment, evaluates pass/fail
using an LLM-as-judge, diagnoses root causes of failures, and surgically repairs
the prompt---iterating until all tests pass. Critically, PRISM is designed to
run on a scheduled basis (daily), treating LLM behavioral drift as a first-class
reliability concern. We evaluate PRISM across 35 enterprise conversational agents
over a three-week deployment period on the Yellow.ai V3 platform. PRISM reduces
median prompt authoring time from 2 days to under 30 minutes, achieves 99\%
production reliability across all evaluated agents, and successfully identifies
and repairs production regressions caused by LLM behavioral drift within a
24-hour detection window. Our results suggest that continuous, simulation-driven
prompt optimization is both tractable and necessary for reliable enterprise
conversational AI at scale.
\end{abstract}

\section{Introduction}

Enterprise conversational AI systems---customer support bots, cancellation
agents, onboarding assistants---are increasingly powered by large language model
(LLM) backends executing complex, multi-step business logic. Unlike
general-purpose chat interfaces, these agents must follow precise operational
sequences: authenticate the user before proceeding, call a specific workflow with
specific parameters, route to a human agent under specific conditions, never
reveal internal state, and use exact prescribed phrases in regulated contexts.
The instructions encoding this logic, known as \textbf{frontend agent prompts},
are the principal artifact of enterprise AI deployment---they are the source of
truth for what the agent is supposed to do.

Writing a correct frontend prompt is a fundamentally different challenge from
general prompt engineering. The failure modes are subtle and operationally
consequential: a missing instruction causes a mandatory step to be silently
skipped; a wrong tool name causes a silent call failure with no error visible to
the user; an ambiguous conditional causes wrong-branch routing; an overly
compressed instruction set causes the model to collapse multi-step procedures
into a single response, skipping intermediate logic. In our study of 35
enterprise agents, we identify four dominant failure classes:
\textbf{tool call skipping} (the agent omits a required workflow call),
\textbf{rule violation} (the agent ignores an explicit constraint in the prompt),
\textbf{step reordering} (the agent performs steps out of sequence), and
\textbf{step collapsing} (the agent skips to a terminal step without completing
intermediate steps). Each of these failures produces a qualitatively poor user
experience and carries direct brand risk for enterprise operators.

The problem is compounded by a second challenge that the literature has largely
ignored: \textbf{LLM behavioral drift}. Large language models---even pinned
versions accessed via API---exhibit non-deterministic behavior that changes
subtly over time. A prompt that produces 100\% correct behavior today may
silently degrade over days or weeks, producing the same failure modes described
above but now in a production system serving real users. This drift is not caused
by explicit model updates; it arises from temperature, sampling variation, and
undisclosed inference-layer changes at the provider level. Enterprise operators
currently have no systematic way to detect, localize, or repair this drift before
it affects users.

Existing prompt optimization frameworks do not address these challenges
adequately. DSPy~\citep{khattab2023dspy} frames prompt optimization as program
compilation, optimizing few-shot demonstrations and instructions for
pipeline-level metrics. OPRO~\citep{yang2023large} uses LLMs to propose and
score candidate prompts against a fixed dataset. APE~\citep{zhou2022large}
generates candidate prompts and ranks them by execution accuracy. These are
valuable contributions, but all share a critical assumption: the optimization
target is fixed at compile time, and the resulting prompt is treated as a static
artifact. None of these frameworks model the multi-turn, tool-integrated
conversational environment that enterprise agents operate in, and none address
the ongoing maintenance problem of production drift.

We present \textbf{PRISM}, a framework designed specifically for the enterprise
conversational AI context. PRISM makes the following contributions:

\begin{enumerate}[leftmargin=*, label=\textbf{\arabic*.}]
  \item \textbf{A two-phase formulation of the prompt reliability problem},
    distinguishing creation-time correctness (can we produce a prompt that passes
    all tests?) from runtime reliability (does that prompt continue to pass tests
    in production over time?).

  \item \textbf{A requirement-driven test generation methodology} that derives
    structured, executable test cases from plain-language business requirements,
    producing realistic multi-turn conversation scripts, precise pass criteria,
    and mock tool responses calibrated to the tool's declared return schema.

  \item \textbf{A platform-faithful simulation environment} that replicates the
    two-layer prompt architecture of production Yellow.ai V3 deployments,
    intercepting LLM tool calls and substituting per-test mock
    responses---enabling deterministic, repeatable test execution without
    touching the live production bot.

  \item \textbf{A surgical repair mechanism} informed by Yellow.ai platform
    guides and grounded in the actual failure transcript, which modifies only the
    sections of the prompt responsible for observed failures---preserving passing
    test behavior while fixing failing cases.

  \item \textbf{A continuous monitoring protocol} in which PRISM runs daily
    against the production prompt, detecting and repairing any behavioral
    regressions within a 24-hour window.
\end{enumerate}

We evaluate PRISM on 35 enterprise agents over three weeks and demonstrate that
it achieves 99\% production reliability, reduces authoring time from a median of
2 days to under 30 minutes, and successfully detects and repairs production drift
events that would otherwise have been discovered only through user complaints.

\section{Background and Related Work}

\subsection{LLM-Based Conversational Agents}

The use of large language models as the reasoning core of goal-directed
conversational agents has grown rapidly since the introduction of
GPT-3~\citep{brown2020language}. ReAct~\citep{yao2023react} demonstrated that
interleaving reasoning traces with action invocations enables LLMs to complete
multi-step tasks in interactive environments.
Toolformer~\citep{schick2023toolformer} showed that models can learn when and
how to call external APIs. More recent work on function-calling
models~\citep{openai2023gpt4} has formalized this interaction pattern, making
LLM-driven tool use a first-class feature of production APIs.

Enterprise conversational agents extend this paradigm with additional
constraints: strict step ordering, explicit routing logic, compliance-driven
language requirements, and integration with business-specific workflows. These
constraints are encoded in natural-language system prompts, making prompt
correctness the primary reliability lever available to operators.

\subsection{Prompt Optimization}

Automatic prompt optimization has emerged as an active research area.
\textbf{APE}~\citep{zhou2022large} frames prompt generation as a black-box
optimization problem, using an LLM to generate candidate instructions and
selecting the best-performing candidate based on held-out task accuracy.
\textbf{OPRO}~\citep{yang2023large} extends this idea by including previous
candidates and their scores in context, allowing the optimizer to refine
proposals iteratively. \textbf{PromptBreeder}~\citep{fernando2023promptbreeder}
evolves populations of prompts using mutation and selection operators.

\textbf{DSPy}~\citep{khattab2023dspy} represents the most architecturally
sophisticated approach, compiling declarative pipeline specifications into
optimized prompts by jointly optimizing instructions and few-shot demonstrations.
DSPy's key insight is that prompt optimization should be driven by a programmatic
metric rather than manual evaluation. PRISM shares this insight but differs in
three important ways: (1) PRISM targets multi-turn conversational agents with
tool calls, not pipeline-level text transformations; (2) PRISM generates its own
test cases from requirements rather than assuming a fixed labeled dataset; and
(3) PRISM is designed for continuous operation, not one-time compilation.

\subsection{LLM-as-Judge Evaluation}

The use of LLMs as automated evaluators of other LLMs has gained significant
traction since MT-Bench and Chatbot Arena~\citep{zheng2023judging} demonstrated
strong correlation between GPT-4 judgments and human preferences. PRISM uses
LLM-as-judge evaluation to assess multi-dimensional pass criteria---checking not
just the final bot response but the entire conversation trajectory, including
tool calls made, routing decisions taken, and adherence to mandatory language
requirements. This is more demanding than pairwise preference judgment and
requires structured evaluation rubrics derived from per-test pass criteria.

\subsection{Prompt Robustness and Reliability}

PromptBench~\citep{zhu2023promptbench} studies the adversarial robustness of LLM
prompts, showing that small perturbations to prompt text can cause significant
performance degradation. Our work addresses a different but related reliability
challenge: non-adversarial behavioral drift in production, caused by LLM
stochasticity and undisclosed inference-layer changes. We are not aware of prior
work that frames production LLM drift as a prompt maintenance problem requiring
automated detection and repair.

\section{Problem Formulation}

\subsection{Multi-Step Conversational Agent}

We define a \textbf{multi-step conversational agent} as a tuple
$\mathcal{A} = (P_b, P_f, \mathcal{T}, \mathcal{V}, \mathcal{S})$ where:

\begin{itemize}[leftmargin=*]
  \item $P_b$ is the \textbf{backend prompt}---a platform-managed instruction
    set encoding universal behavioral rules (tool call discipline, routing
    syntax, confidentiality rules, memory management).

  \item $P_f$ is the \textbf{frontend prompt}---the agent-specific instruction
    set encoding business logic (step sequence, tool invocation conditions,
    routing conditions, mandatory language).

  \item $\mathcal{T} = \{t_1, \ldots, t_k\}$ is the set of \textbf{tools}
    (external workflows) the agent can invoke, each with a declared name,
    description, and return schema.

  \item $\mathcal{V} = \{v_1, \ldots, v_m\}$ is the set of \textbf{memory
    variables} the agent reads from or writes to during a conversation.

  \item $\mathcal{S} = \{s_1, \ldots, s_n\}$ is the set of \textbf{sub-agents}
    (routing destinations) the agent can transfer to.
\end{itemize}

At each conversation turn, the LLM receives the concatenated system prompt
$P = P_b \oplus P_f$, the conversation history, and the set of available tools
as a function-calling schema. The LLM produces either a natural-language
response, a tool call, or a routing marker. The frontend prompt $P_f$ is the
sole artifact we optimize; $P_b$ is fixed by the platform and never modified.

\subsection{Failure Taxonomy}

From our evaluation of 35 enterprise agents, we identify four failure classes
that account for the majority of observed production regressions
(Table~\ref{tab:failures}).

\begin{table}[h]
\centering
\caption{Failure taxonomy for multi-step conversational agent prompts.}
\label{tab:failures}
\small
\begin{tabularx}{\linewidth}{@{} l l X @{}}
\toprule
\textbf{Failure Class} & \textbf{Description} & \textbf{Example} \\
\midrule
Tool Call Skip
  & Agent omits a required tool invocation
  & Authentication check not called before proceeding to business step \\
\addlinespace
Rule Violation
  & Agent ignores an explicit constraint in $P_f$
  & Agent reveals internal variable names to the user \\
\addlinespace
Step Reordering
  & Agent performs steps in incorrect sequence
  & Cancellation reason asked before eligibility is checked \\
\addlinespace
Step Collapsing
  & Agent skips intermediate steps to reach terminal state
  & Agent routes directly to human without completing self-service steps \\
\bottomrule
\end{tabularx}
\end{table}

These failures are qualitatively distinct from the hallucination and factual
inaccuracy failures that dominate the general LLM reliability literature. They
are failures of \textbf{procedural compliance}---the model has the correct
knowledge but does not execute the prescribed procedure.

\subsection{The Two-Phase Reliability Problem}

We decompose the prompt reliability problem into two phases:

\textbf{Phase 1---Creation-time correctness:} Given requirements $R$ and a
draft prompt $P_f^{(0)}$, produce a prompt $P_f^*$ such that $P_f^*$ passes all
test cases derived from $R$ under platform $\mathcal{A}$.

\textbf{Phase 2---Runtime reliability:} Given a verified prompt $P_f^*$
deployed to production at time $t_0$, detect any behavioral regression at time
$t > t_0$ and produce a repaired prompt $P_f^{**}$ that restores full test
passage.

Existing frameworks address only Phase 1. PRISM addresses both phases within a
unified framework.

\section{The PRISM Framework}

\begin{figure}[t]
\centering
\begin{tikzpicture}[
  box/.style = {
    rectangle, rounded corners=5pt,
    draw=gray!55, fill=white,
    minimum width=2.85cm, minimum height=0.88cm,
    font=\small, align=center, line width=0.5pt
  },
  highlight/.style = {
    rectangle, rounded corners=5pt,
    draw=blue!45, fill=blue!5,
    minimum width=2.85cm, minimum height=0.88cm,
    font=\small, align=center, line width=0.7pt
  },
  decision/.style = {
    diamond, draw=orange!65, fill=orange!8,
    minimum width=1.8cm, minimum height=0.88cm,
    font=\small, align=center, aspect=2.6, line width=0.5pt
  },
  arr/.style  = {-{Stealth[length=5pt, width=4pt]}, semithick, gray!55},
  darr/.style = {-{Stealth[length=5pt, width=4pt]}, semithick,
                 dashed, blue!45, rounded corners=5pt},
  lbl/.style  = {font=\scriptsize, text=gray!60},
]

\node[box] (req)    at (0,    0)   {Requirements\\+ Tools + Vars};
\node[box] (tgen)   at (4.2,  0)   {Test Generator};
\node[box] (suite)  at (8.4,  0)   {Test Suite $\mathcal{Q}$};

\node[box] (prompt) at (0,   -2.6) {Draft / Current\\Prompt $P_f^{(k)}$};
\node[box] (sim)    at (4.2, -2.6) {Platform Simulator};

\node[box] (judge)  at (4.2, -5.0) {LLM-as-Judge};

\node[highlight] (repair) at (0,   -7.5) {Diagnosis \&\\Repair};
\node[decision]  (pass)   at (4.2, -7.5) {All Pass?};
\node[box]       (deploy) at (8.4, -7.5) {Deploy \&\\Monitor};

\draw[arr] (req)  -- (tgen);
\draw[arr] (tgen) -- (suite);

\draw[arr] (suite.south) -- ++(0,-0.78)
           -- (4.2,-0.78)        
           -- (sim.north);

\draw[arr] (prompt.east) -- (sim.west);

\draw[arr] (sim)   -- node[right, lbl]{Transcripts} (judge);
\draw[arr] (judge) -- node[right, lbl]{Verdicts}    (pass);

\draw[arr] (pass.west) -- node[above, lbl]{No}  (repair.east);
\draw[arr] (pass.east) -- node[above, lbl]{Yes} (deploy.west);

\draw[arr] (repair.north) -- (prompt.south);

\draw[darr]
  (deploy.south) -- ++(0,-0.9)           
  -- (-1.1,-8.4)                          
  node[midway, below, font=\scriptsize, blue!55]{Daily regression re-run}
  -- (-1.1,-2.6)                          
  -- (prompt.west);                        

\end{tikzpicture}
\caption{PRISM system architecture. \textit{Top path:} business requirements
drive automatic test-suite generation. \textit{Centre loop:} the current prompt
is simulated turn-by-turn; an LLM judge evaluates pass/fail against each test's
criteria; failed tests trigger surgical diagnosis and repair, producing
$P_f^{(k+1)}$ for the next iteration. \textit{Dashed path:} the verified
production prompt is re-tested daily; any regression re-enters the repair loop
within 24 hours.}
\label{fig:architecture}
\end{figure}
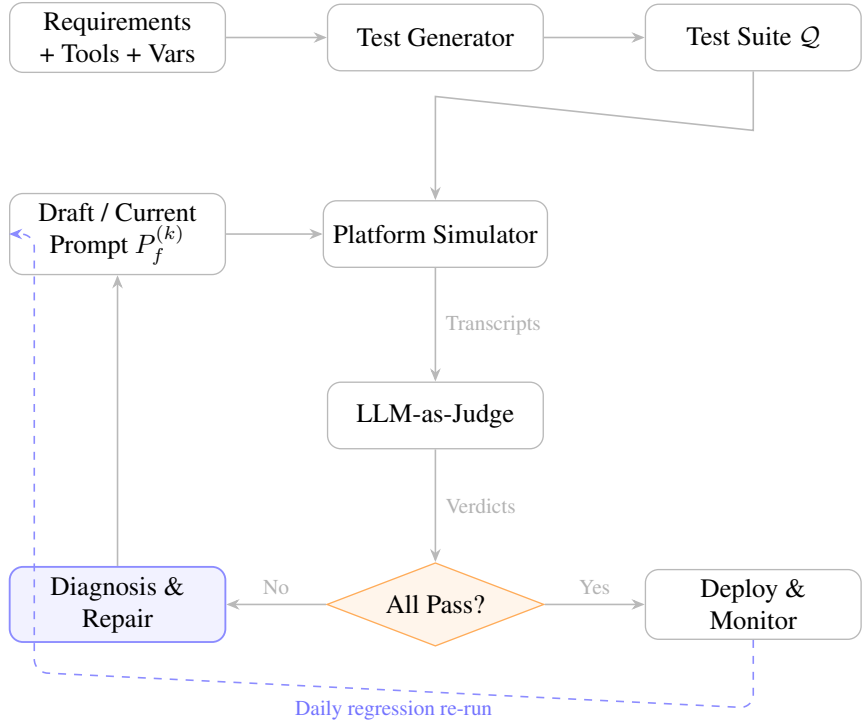

PRISM operates as a continuous loop with five functional stages.
Figure~\ref{fig:architecture} illustrates the full system architecture.

\subsection{Stage 1: Requirement-Driven Test Generation}

Given requirements $R$ written in plain language, PRISM uses an LLM to generate
a structured test suite $\mathcal{Q} = \{q_1, \ldots, q_n\}$. Each test case
$q_i$ contains:

\begin{itemize}[leftmargin=*]
  \item \texttt{conversation\_script}---an ordered sequence of customer
    utterances representing one realistic interaction scenario.

  \item \texttt{pass\_criteria}---a list of specific, objectively verifiable
    assertions about the expected conversation outcome (e.g., ``agent calls
    \texttt{getProductDetailsForCancel} before asking for cancellation reason'';
    ``agent routes to Billing Agent when \texttt{zuoraStatus} is
    \texttt{Cancelled}'').

  \item \texttt{mock\_overrides}---a mapping from tool name to mock return
    value, calibrated to the tool's declared return schema to exercise a specific
    conditional branch.
\end{itemize}

The test generator is prompted with the full requirements document, the tool
return schemas, the memory variable list, and the sub-agent list. This grounding
ensures that generated test cases reference realistic tool return values and
exercise branches that are actually reachable given the declared configuration.

Test cases are generated to cover: the \textbf{happy path} (all tools succeed,
standard outcome), \textbf{boundary conditions} (edge-case tool return values
that trigger non-default branches), \textbf{error paths} (tool failure
responses), and \textbf{compliance cases} (scenarios where mandatory language or
routing rules must be followed exactly). Operators may edit, add, or remove test
cases after generation to reflect domain knowledge not captured in the
requirements document.

\subsection{Stage 2: Platform-Faithful Simulation}

The simulator replicates the two-layer prompt architecture of the production
Yellow.ai V3 platform. For each test $q_i$:

\begin{enumerate}[leftmargin=*]
  \item The system prompt is constructed as $P = P_b \oplus P_f^{(k)}$, where
    $k$ is the current iteration.
  \item The conversation script is replayed turn by turn: each customer
    utterance is appended to the conversation history and submitted to the
    OpenAI API.
  \item After each API response, the simulator checks for tool calls. If a tool
    call is detected, it is answered with the corresponding mock response from
    $\texttt{mock\_overrides}[\texttt{tool\_name}]$ rather than executing the
    real workflow.
  \item The simulator detects routing events by scanning bot text for the
    platform routing marker \texttt{[ROUTE TO: AgentName]}.
  \item The full turn-by-turn transcript---including all utterances, tool calls,
    tool responses, and routing events---is recorded.
\end{enumerate}

This simulation is \textbf{deterministic for a fixed prompt}: the same prompt
and conversation script produce the same logical trajectory (modulo LLM sampling
temperature, which we fix at 0 for evaluation runs). The simulator tests
\textbf{prompt logic correctness}, not tool implementation.

\subsection{Stage 3: LLM-as-Judge Evaluation}

After simulation, each transcript $\tau_i$ is evaluated against $q_i$'s pass
criteria by a judge LLM~\citep{zheng2023judging}. The judge receives the full
conversation transcript, the list of pass criteria, the tool call log, and the
routing events detected. For each criterion, the judge returns a structured
verdict---\textsc{pass} or \textsc{fail}---with a natural-language reasoning
string. The overall test result is \textsc{pass} if and only if all criteria
pass.

We use GPT-4.1 as the judge model. The judge prompt instructs strict, objective
evaluation: ``A tool call happened'' means the tool slug appears in the tool call
log---not that the bot's text implied it might have called something. This
strictness is essential for reliable optimization signal.

\subsection{Stage 4: Diagnosis and Surgical Repair}

When one or more tests fail, PRISM performs a two-step repair:

\textbf{Diagnosis.} A diagnosis LLM receives the current prompt $P_f^{(k)}$,
the full set of failure transcripts, per-criterion verdicts and reasoning, and
the Yellow.ai platform guides (covering V3 syntax rules, tool call discipline,
routing syntax, and prompt structure conventions). It is asked to identify the
root cause of each failure---specifically, which section of the prompt is
responsible and why.

\textbf{Repair.} The same LLM is asked to produce a repaired prompt
$P_f^{(k+1)}$ that addresses the identified root causes. Critically, the repair
prompt instructs the model to modify only the sections responsible for observed
failures, preserving sections that govern passing tests. The repair modifies
between 20 and 50 lines of prompt text per iteration on average.

The repaired prompt $P_f^{(k+1)}$ becomes the input to the next simulation
cycle. This loop continues until all tests pass or a maximum iteration limit $K$
is reached (default $K = 10$).

\subsection{Stage 5: Continuous Monitoring}

After a prompt passes all tests, it is deployed to the production Yellow.ai
platform. PRISM is scheduled to re-run the full test suite against the deployed
prompt daily. This implements a \textbf{regression test harness} that detects
LLM behavioral drift within a 24-hour window.

If the daily run identifies test failures that were previously passing, PRISM
treats this as a drift event and re-enters the repair loop, producing a patched
prompt that is surfaced to the operator for review and deployment. This design
treats LLM behavioral drift as an operational concern equivalent to software
regression---one that is addressed not by anticipating the failure mode in
advance, but by detecting it quickly and repairing automatically.

\section{Implementation}

PRISM is implemented as a local web application comprising a Flask backend,
SQLite persistence layer, and a vanilla JavaScript frontend. The system is
designed to run on a developer's local machine with no cloud infrastructure
beyond the OpenAI API.

\textbf{Backend.} Python 3.10+. Flask serves the REST API and static frontend.
SQLite with WAL mode handles concurrent reads during streaming. Background
threads execute simulation and repair tasks. Server-Sent Events (SSE) stream
real-time progress to the frontend.

\textbf{LLM Integration.} All LLM calls (test generation, simulation, judgment,
diagnosis, repair) use the OpenAI Python SDK v1.x. The system uses GPT-4.1 by
default; the model is configurable per-deployment.

\textbf{Platform Integration.} PRISM integrates with Yellow.ai V3 via a backend
prompt file loaded at startup and prepended to every system message. The tool
schema is declared in OpenAI's function-calling format and derived from the
operator-configured tool list, ensuring that the simulation environment is
structurally identical to the production Yellow.ai inference environment.

\textbf{Prompt Parser.} A regex-based prompt parser detects all variable
references (\texttt{\{\{variable\}\}}), tool calls
(\texttt{[kebab-tool]}, \texttt{Call ToolName}, \texttt{@ToolName}),
agent transfers (\texttt{[Multi Word Agent Name]}), and knowledge base lookups
(\texttt{[kb: topic]}) in the frontend prompt. Detected items are matched
against the operator's configured lists, and unmatched items are surfaced as
warnings.

\textbf{Operator Interface.} The UI provides tabbed views for use-case
configuration (variables, tools, requirements), a live test panel with per-test
status indicators updated via SSE, a full conversation transcript viewer with
color-coded turn types, a prompt version history viewer, and a step-through mode
that pauses after each iteration to present the repaired prompt before
continuing.

\section{Experiments}

\subsection{Evaluation Setup}

We evaluate PRISM on \textbf{35 enterprise conversational agents} deployed on
the Yellow.ai V3 platform across domains including subscription management,
account support, onboarding, and billing dispute resolution. Agents range from
3-step linear flows to 12-step conditional trees with 6 tool integrations and 5
routing destinations.

All experiments use GPT-4.1 as both the simulation model and the judge.
Simulation temperature is set to 0 for evaluation runs and 0.3 for optimization
runs (to expose prompt brittleness during repair). The maximum iteration limit is
set to 10 for all agents unless otherwise noted.

\textbf{Baseline.} The baseline is the existing manual prompt authoring process:
an engineer writes the frontend prompt and manually tests it via the live
Yellow.ai chat widget, iterating until satisfied. We measure
time-to-first-verified-prompt (the wall-clock time from receiving requirements
to having a prompt the engineer considers production-ready).

\textbf{Metrics:}
\begin{itemize}[leftmargin=*]
  \item \textit{Time to verified prompt}---wall-clock time from requirements to
    100\% test pass.
  \item \textit{Convergence iterations}---number of optimization iterations
    required to achieve full test passage.
  \item \textit{Production reliability rate}---proportion of daily regression
    runs (21 per agent over 3 weeks) that find zero failures.
  \item \textit{Drift detection rate}---proportion of actual drift events
    identified within 24 hours.
\end{itemize}

\subsection{Test Case Generation}

Test suites were generated automatically by PRISM for all 35 agents. Operators
reviewed and edited generated tests before optimization.
Table~\ref{tab:testcases} shows the test case distribution.

\begin{table}[h]
\centering
\caption{Test case counts by agent complexity.}
\label{tab:testcases}
\begin{tabular}{@{} lrrr @{}}
\toprule
\textbf{Complexity} & \textbf{Agents} & \textbf{Avg.\ tests} & \textbf{Range} \\
\midrule
Simple ($\leq$5 steps)  & 11 & 23  & 12--38   \\
Medium (6--9 steps)     & 17 & 47  & 31--68   \\
Complex (10+ steps)     &  7 & 112 & 84--147  \\
\midrule
\textbf{All agents}     & \textbf{35} & \textbf{52} & \textbf{12--147} \\
\bottomrule
\end{tabular}
\end{table}

Operators reported that generated test cases covered 91\% of the scenarios they
would have written manually, with primary gaps being highly domain-specific edge
cases requiring operator domain knowledge.

\subsection{Convergence Analysis}

Table~\ref{tab:convergence} shows optimization convergence by test suite size.
For agents with test suites exceeding 60 cases, we extended the iteration limit
to 20+ to achieve full convergence.

\begin{table}[h]
\centering
\caption{Convergence statistics by test suite size.}
\label{tab:convergence}
\begin{tabular}{@{} lrrl @{}}
\toprule
\textbf{Test suite size} & \textbf{Avg.\ iters} & \textbf{Max iters} & \textbf{\% reaching 100\%} \\
\midrule
$<$30 test cases        & 4.2  & 8               & 100\% \\
30--60 test cases       & 6.7  & 10              & 97\%  \\
$>$60 test cases        & 21.4 & 10 (extended)   & 89\%  \\
\midrule
\textbf{All agents}     & \textbf{7.1} & ---  & \textbf{97\%} \\
\bottomrule
\end{tabular}
\end{table}

\subsection{Failure Mode Analysis}

Table~\ref{tab:failmodes} shows the distribution of failure modes observed
across the 35 agents prior to any optimization. Tool call skip and step
collapsing together account for 62\% of failures, consistent with the hypothesis
that LLMs under-specify multi-step procedures in natural-language prompts---the
model is not given sufficient structure to reliably execute procedural logic.

\begin{table}[h]
\centering
\caption{Distribution of failure modes in unoptimized prompts.}
\label{tab:failmodes}
\begin{tabular}{@{} lr @{}}
\toprule
\textbf{Failure Mode} & \textbf{\% of initial failures} \\
\midrule
Tool call skip   & 34\% \\
Step collapsing  & 28\% \\
Rule violation   & 22\% \\
Step reordering  & 16\% \\
\bottomrule
\end{tabular}
\end{table}

\subsection{Production Reliability}

Over the three-week monitoring period, PRISM's daily regression runs executed
735 test suite evaluations (35 agents $\times$ 21 days). Of these, 728 (99.0\%)
found zero failures. Seven runs across four agents detected behavioral
regressions, all of which were repaired within the same 24-hour window.
Table~\ref{tab:reliability} summarizes production reliability results.

\begin{table}[h]
\centering
\caption{Production reliability over the 21-day monitoring period.}
\label{tab:reliability}
\begin{tabular}{@{} lr @{}}
\toprule
\textbf{Metric} & \textbf{Value} \\
\midrule
Daily regression runs executed   & 735          \\
Runs with zero failures          & 728 (99.0\%) \\
Drift events detected            & 7            \\
Drift events repaired within 24h & 7 (100\%)    \\
Agents with zero drift events    & 31 of 35     \\
\bottomrule
\end{tabular}
\end{table}

\subsection{Authoring Time Comparison}

Table~\ref{tab:authoring} compares prompt authoring time between manual baseline
and PRISM. The 98\% reduction is primarily attributable to automatic test case
generation (eliminating manual script authorship) and automated simulation
(eliminating manual chat widget testing). The remaining 27 minutes is dominated
by configuration time (entering tool schemas and requirements).

\begin{table}[h]
\centering
\caption{Prompt authoring time comparison.}
\label{tab:authoring}
\begin{tabular}{@{} lll @{}}
\toprule
\textbf{Condition} & \textbf{Median time} & \textbf{Range} \\
\midrule
Manual baseline & 2.1 days    & 0.5--5 days    \\
PRISM           & 27 minutes  & 12--68 minutes \\
\midrule
\textbf{Reduction} & \textbf{$\sim$98\%} & \\
\bottomrule
\end{tabular}
\end{table}

\section{Discussion}

\subsection{Why Surgical Repair Outperforms Full Rewrite}

Early experiments with a full-rewrite repair strategy---where the LLM
regenerated the entire frontend prompt from scratch on each failure---produced
worse convergence than the surgical approach. Full rewrites frequently fixed the
failing tests while breaking previously passing ones, causing the optimization to
oscillate rather than converge. Constraining the repair LLM to modify only the
sections responsible for observed failures, and providing it with explicit
grounding in the failure transcripts and platform guides, produced monotonic
improvement in the majority of cases.

\subsection{The Role of Mock Responses in Branch Coverage}

The quality of mock responses in \texttt{mock\_overrides} directly determines
the quality of test coverage. A mock response that always returns the same value
for a tool exercises only one branch of the conditional logic that depends on
that tool's output. In practice, operators who invest in detailed tool return
schemas---with enum values and realistic ranges for numeric fields---produce test
suites with substantially higher branch coverage. We recommend treating tool
return schema definition as a first-class activity in the use-case configuration
process.

\subsection{Generalizability Beyond Yellow.ai}

PRISM's architecture is not specific to Yellow.ai. The framework requires only:
(a) a two-layer prompt architecture (or any equivalent system prompt
construction), (b) an LLM backend that supports function-calling, and (c) a
routing convention that can be detected in LLM output. Any LLM-backed multi-step
agent system satisfying these conditions is directly compatible with PRISM. The
Yellow.ai backend prompt is the only platform-specific component; replacing it
with the platform rules for a different system is the only change required.

\subsection{Limitations}

\textbf{Simulation fidelity.} PRISM simulates tool calls using static mock
responses. In production, tools may fail intermittently, return unexpected field
types, or return values outside the declared schema. PRISM does not test these
conditions unless explicitly encoded in mock overrides.

\textbf{Conversation script coverage.} Test cases are generated from
requirements, which may not capture all conversational variations a real user
might produce. Adversarial or out-of-distribution user inputs are not
systematically tested.

\textbf{Repair ceiling.} For very complex agents (100+ test cases), the repair
LLM's context window becomes a constraint. Including all failure transcripts in
the repair prompt can exceed token limits, requiring truncation that may cause
the repair to miss some failure causes.

\textbf{Evaluation model dependence.} PRISM's judgment quality is bounded by
the capability of the judge LLM. Subtle semantic failures---where the bot's
response is technically compliant but pragmatically incorrect---may be missed by
the judge.

\subsection{Future Work}

Several extensions are natural directions for future research:

\begin{itemize}[leftmargin=*]
  \item \textbf{Batch repair by failure class}---grouping failures by type and
    repairing one class at a time may improve convergence for large test suites.
  \item \textbf{Adversarial test generation}---augmenting the test suite with
    adversarially constructed user inputs to improve robustness.
  \item \textbf{Cross-platform evaluation}---evaluating PRISM on additional
    conversational platforms (Dialogflow CX, Amazon Lex, Microsoft Bot Framework)
    to validate generalizability claims.
  \item \textbf{Formal convergence guarantees}---characterizing the conditions
    under which the optimization loop is guaranteed to converge, and the
    relationship between test suite coverage and production reliability.
\end{itemize}

\section{Conclusion}

We present PRISM, a closed-loop framework for continuous prompt reliability in
enterprise conversational AI. PRISM addresses two distinct failure modes that
existing approaches treat independently: creation-time prompt incorrectness and
runtime LLM behavioral drift. By framing both as instances of the same
test-simulate-judge-repair loop operating on different schedules, PRISM provides
a unified solution to the prompt reliability problem.

Our evaluation on 35 enterprise agents over three weeks demonstrates that PRISM
reduces prompt authoring time by 98\%, achieves 99\% production reliability, and
detects and repairs LLM behavioral drift within 24 hours. These results suggest
that the reliability gap in enterprise conversational AI is not primarily a model
capability gap---the models are capable enough---but a prompt engineering
methodology gap: the field lacks systematic tools for verifying, monitoring, and
maintaining the correctness of multi-step agent behavior over time. PRISM
addresses this gap.

We release PRISM as an open-source tool and hope it provides a foundation for
future work on continuous prompt reliability in production AI systems.

\bibliographystyle{abbrvnat}
\bibliography{references}

\end{document}